\documentclass[12pt]{alt2022} 

\usepackage{booktabs}       
\usepackage{amsfonts}       
\usepackage{bbm}
\usepackage{amssymb}
\usepackage{algorithm}
\usepackage{cleveref}

\crefname{lemma}{lemma}{lemmas}

\title[Empirical Monte Carlo Bounds]{Empirical Error Bounds for Monte Carlo Search}
\usepackage{times}

\altauthor{%
    \Name{John Mern} \Email{jmern91@stanford.edu} \\ 
    \Name{Mykel J. Kochenderfer} \Email{mykel@stanford.edu} \\
    \addr Stanford, CA, USA%
 }

\begin{document}

\maketitle

\begin{abstract}%
The theoretical asymptotic bounds provided for many Monte Carlo methods cannot be calculated using known quantities during search. 
Often search results provide no principled measure of confidence and may be sub-optimal due to premature termination.
In this work, we prove two sets of bounds on the error rate of Monte Carlo search over non-stationary bandits and Markov decision processes. 
The presented bounds hold for general Monte Carlo solvers meeting mild convergence conditions.
These bounds can be directly computed at the conclusion of the search and do not require knowledge of the true action-value, allowing them to be used as search stopping criteria. 
We also provide a simple sub-optimality probability estimation method based on the presented bounds. 
We empirically test the tightness of the bounds and accuracy of the estimator through experiments on a multi-armed bandit and a discrete Markov decision process.
\end{abstract}


\section{Introduction}
Monte Carlo (MC) estimation can be used to solve decision making problems when the dynamics of the environment are not fully known. 
Solvers draw samples from a generative model of the environment to estimate the values of different actions. 
Typically, solvers that use Monte Carlo estimation sample actions during simulation according to a non-stationary search policy.
In these cases, empirical error bounds cannot be computed using standard methods that assume independent identically distributed ($iid$) samples.
Because of this, many solvers return action value estimates without a confidence measure. 



Solvers based on Monte Carlo sampling have been proposed for single-step and sequential decision problems~\citep{sutton1998}.
Online search methods such as Monte Carlo tree search (MCTS) have solved a variety of complex sequential decision making tasks~\citep{silver2016, silver2017}, though they are computationally expensive to execute.
The expected error of a Monte Carlo estimate 
tends to decrease with the number of samples drawn~\citep{audibert2009}. 
When expected error cannot be computed during search, arbitrary, heuristic stopping conditions are often used.
This can lead to high computation cost from excessive sampling or premature termination and sub-optimal results.
Measurable bounds or estimates of error probability would provide a more useful stopping criteria.

Several Monte Carlo solvers have theoretical asymptotic bounds on measures such as regret~\citep, PAC-correctness~\citep{kearns1999, kaufmann2017}, and error probability~\citep{shah2020}.
These theoretical bounds are defined in terms of parameters not generally known during search. 
For example, regret bounds for the upper confidence tree (UCT) MCTS algorithm are defined in terms of the difference between the converged Monte Carlo estimate and the true optimal discounted return~\citep{kocsis2006}. 
These theoretical expressions help to understand algorithm behavior, but they cannot be numerically calculated.
Empirical bounds may be calculated using only values observed from simulation and can be used to evaluate solver performance during search.
Empirical bounds have been proven for some MC solvers under special circumstances~\citep{grill2016, shah2020}, but are not generally available.

\subsection{Our Contributions}
In this work, we present two sets of finite-sample, empirical bounds on error probability for general Monte Carlo solvers. 
One set of bounds can be applied to any Monte Carlo estimate of problem with bounded value.
The second set of bounds provides tighter limits for problems in which the central limit theorem (CLT) can be applied. 
In both cases, we present bounds on the probability that the action value estimated by the search overestimates the true value by more than $\epsilon$ as
\begin{equation}
    P(\bar{Q}(s, a) \geq Q(s, a) + \epsilon) \leq \alpha
\end{equation}
where $Q(s, a)$ is the expected sum of discounted returns from taking action $a$ from state $s$ and $\bar{Q}(s, a)$ is the action value estimate from the MC search. 
We also bound the probability that the action returned from search $a$ is worse than any other considered action $a'$ by more than $\epsilon$ as
\begin{equation}
    P(Q(s, a) \leq Q(s, a') - \epsilon) \leq \alpha
\end{equation} 
when following the search policy for the remaining trajectory. 
Both sets of bounds can be numerically calculated during search for use as a confidence measure or as a stopping criteria. 
We also present terms to estimate both of the presented error probabilities without upper-bound guarantees. 

We evaluated the tightness of the bounds and the accuracy of the estimation through numerical experiments.
We tested performance on a multi-armed bandit and a discrete action space navigation task.
A simple, fixed-depth Monte Carlo solver was used for evaluation on both tasks. 
Additionally, UCT-MCTS was tested on the sequential navigation task. 
The general Monte Carlo bounds hold for all tested configurations, and the CLT bounds hold when sufficient samples are drawn. 
Bound gaps tended to decrease with increased sample set size. 
The error estimate tends to produce a closer approximation of the actual error rates than the bounds.

\subsection{Related Work}
Multi-armed bandit problems have been extensively studied by the statistics and learning communities~\citep{lai1985}.
Prior work in statistical analysis of random processes provide foundational measures toward generally applicable non-asymptotic bounds.
Prior works develop empirical Bernstein-type bounds based on sample variance~\citep{maurer2009}.
Early work proved bounds for sums of independent, identically distributed random random variables based on sample mean and variance~\citep{beygelzimer2011}.
Further work extended these ideas to functions over martingales~\citep{peel2013, zhang2021} with some restrictions. 

Many algorithms have been proposed to approximately solve the bandit problem. 
Foundational algorithms, such as UCB1, have been extensively studied~\citep{auer2002}.
Similarly, many MCTS methods have been proposed to solve MDPs and partially observable MDPs (POMDPs).
UCT is a well-studied algorithm~\citep{kocsis2006} that has been used as the basis for several variants~\citep{silver2010}.
Convergence guarantees and non-asymptotic performance bounds have been proved for UCT and other Monte Carlo methods.
These bounds are often theoretical in that they are defined in terms of the unknown parameters such as the true expected return value. 
Few algorithms provide performance bounds that can be numerically calculated using only empirically observed values. 

Some solvers have been proposed with calculable non-asymptotic bounds.
\citet{shah2020} present non-asymptotic analysis of UCT with improved regret-bounds, and additionally propose a modified, fixed-depth tree search with policy-improvement guarantees.
Early work on approximate sampling by~\citet{kearns1999} proved finite-sample convergence bounds for fixed-depth search, with limited scalability to large problem spaces. 
An MCTS approach to simple regret minimization with finite-sample guarantees was proposed by~\citet{feldman2014}. 
Several other MCTS approaches have been proposed that provide finite-sample error guarantees~\citep{grill2016, kaufmann2017, grill2019}. 
While the bounds in these works hold for their respective algorithms, they are not generally applicable to other search methods. 


\section{Preliminaries}\label{sec: Prelim}
Monte Carlo search estimates the values of taking actions from a given state. 
The value of taking action $a$ in state $s$ and then following policy $\pi$ is denoted $Q^\pi(s,a)$. 
Values associated with the optimal policy are denoted $Q^*(s,a)$. 
Monte Carlo search is used as a policy by repeatedly searching from each encountered state and taking the action with the highest estimated value. 
We will denote values associated with a Monte Carlo policy with superscript $\pi$.
For time-discounted problems, the time-discount rate is denoted $\gamma$.

The set of all available actions is denoted $\mathcal{A}$ and the set of all actions included in a Monte Carlo search is $\mathcal{A}_T \subseteq \mathcal{A}$.
For ab action $a_i$ the value is estimated using the sample set $\mathcal{C}_i \gets (q_i^{(1)}, \dots, q_i^{(n)})$.
The value estimate of action $a_i$ is then $\bar{Q}^{(n)}_i = \frac{1}{n_i}\sum_{q \in \mathcal{C}_i} q$, where $n_i = |\mathcal{C}_i|$.
The notation $\bar{X}^{(n)}$ denotes the Monte Carlo estimate of true quantity $X$ with $n$ samples.
The sample variance of the estimator is denoted $V_i$.

The bounds presented in this work may be applied to Monte Carlo search over sequential decision problems with non-stationary dynamics. 
This includes multi-armed bandits (MABs) with non-stationary payout distributions and Markov decision processes (MDPs). 
The bounds may be applied to Monte Carlo estimates over problems satisfying the following conditions almost surely. 
\begin{enumerate}
    \item Bounded Returns: The sum of discounted rewards is bounded by some non-zero constant $a$ almost surely as $|\bar{Q}^{(n)}_i| \leq b \ \forall \ i, n$.
    \item Convergent Search: The estimated action values converge to the value of the search policy with infinite samples as $\mathbb{E}[\bar{Q}(s, a)] = Q^\pi(s,a)$.  
    \item Accurate Simulation: The generative model used for simulation exactly matches the true environment dynamics.
\end{enumerate}
For MDPs, we additionally assume that the search simulates trajectories from a given starting state $s$ to some potentially variable depth $h$ and uses a given estimator for the value the remaining trajectory.
We refer to the bounds that only require satisfying these conditions as \emph{general} bounds.

We propose an additional set of bounds for problems to which the central limit theorem applies. 
We refer to these bounds as CLT bounds. 
In addition to the above conditions, application of these bounds assumes that distributions of MC estimates approaches a normal distribution with infinite samples as 
\begin{equation}
    \sqrt{n}\big(\bar{Q}(s,a) - Q(s,a)\big) \xrightarrow{d} \mathcal{N}(0, \sigma^2)
\end{equation}
where $\sigma^2$ is the variance of returns under the optimal policy. 
For samples generated under a stationary policy, the CLT should hold. 
Early works by~\citet{markov1951}, \citet{Bernstein1927}, and~\citet{dobrushin1956} proved that the CLT holds for payoffs from non-stationary Markov chains meeting certain ergodic conditions.
Proving the CLT for additional cases is an active area of research~\citep{sinn2013, arlotto2016}.
These may be extended to return estimates under non-stationary policies, however, this is left for future work.

\section{Empirical Bounds and Estimator}
We present two sets of error bounds for Monte Carlo search.
The first set of general bounds apply to search over any problem meeting the conditions in~\cref{sec: Prelim}. 
The second set of bounds are only valid for problems in which the central limit theorem holds. 
We also propose terms to estimate the errors presented without enforcing an upper bound. 

\subsection{General Monte Carlo Bound}

We present bounds on two types of errors that may result from a Monte Carlo search returning an action $a$ with estimated optimal value $\bar{Q}^*(s,a)$.  
The first is a bound on the probability that a value is over-estimated by the Monte Carlo value.
We refer to this as an \emph{value error} bound.
The second is a bound on the probability that an action $a' \in \mathcal{A}_T/a$ is better than the action recommended by search.
We refer to this as a bound on the \emph{action error} probability.

The value error bound gives the probability that an estimated action value is over-estimates the optimal action value.
It is stated formally in~\cref{thm: gen as bound}.
\begin{theorem}~\label{thm: gen as bound}
For a Monte Carlo search returning action value estimate $\bar{Q}_i = q_i$, the probability that the true optimal action value is overestimated $\bar{Q}_i - Q^*_i \geq \epsilon$ is no greater than
\begin{equation}~\label{eq: sub bound}
\exp\Big(\frac{-n_i(|\epsilon - \zeta_i|^+)^2}{2\hat{\sigma}^{2, \alpha}_i}\Big) + \alpha
\end{equation}
where $n_i$ is the number of sampled trajectories in $\mathcal{C}_i$.
An upper bound on the return variance is given by $\hat{\sigma}^{2, \alpha}$ with significance $\alpha \in (0, 1)$.
The $\zeta$ term is an upper bound on the expected bias of the baseline value estimate.
Here, $|x|^+$ represents the $\max(0, x)$ function.
\end{theorem}

The variance upper bound in~\cref{eq: sub bound} is defined for bounded random variables as
\begin{equation}~\label{eq: var bound}
    \hat{\sigma}^{2, \alpha} \gets b^2\Bigg(\frac{1}{b}V_n^\frac{1}{2} + \sqrt{\frac{-\ln{\alpha}}{n - 1}}  \Bigg)^2
\end{equation}
where $V_n$ is the sample variance over $n$ sampled trajectory returns and $b$ is the range of the sample values.
Taking the minimum of~\cref{eq: sub bound} over $\alpha$ gives the tightest bound.
In the bandit setting, the bias term $\zeta = 0$, since the values are sampled directly from the true distributions. 
In the MDP setting, the $\zeta$ term is an upper bound on the bias, defined as 
\begin{equation}
    \zeta = \frac{1}{|\mathcal{C}|}\sum_{k \in \mathcal{C}}\gamma^{H_k} \epsilon^U_k 
\end{equation}
where $H_k$ is the depth of the Monte Carlo rollout, $\gamma \in [0, 1]$ is a discount factor, and $\epsilon_k^U$ is an upper bound on the error of the baseline value estimate at the end of the trajectory. 
At worst, $\epsilon_k^U \gets \max(Q_{max} - \hat{Q}_k, \hat{Q}_k - Q_{min})$, where $\hat{Q}_k$ is the value of the baseline estimate at the end of the trajectory $k$.
The error is zero for trajectories reaching termination.

The presented bound is entirely empirical and requires no unknown values to compute.
It can be used as a stopping criteria to ensure a policy meets a return target with high probability. 
As can be seen, the probability decays exponentially with the number of trajectories sampled for the given action. 
For MDPs, the search-bias decreases with the depth $H$ of each trajectory sample as $\gamma^H$. 

The action error bound limits the probability that an action from the search set is better than the recommended action when the Monte Carlo search policy is followed for subsequent steps.
It is stated formally in~\cref{thm: gen ae bound}.
\begin{corollary}~\label{thm: gen ae bound}
Let a Monte Carlo search return an action $a_i$ with estimated value $\bar{Q}_i$.
For any other action $a_{j} \in \mathcal{A}_T/a_i$ with value estimate $\bar{Q}_j = \bar{Q}_i - \delta$, the probability that the true value of $a_j$ is $Q^\pi_j \geq Q^\pi_i + \epsilon$ is no greater than
\begin{equation}~\label{eq: gen ae bound}
\exp\Big(\frac{-n_i n_j(|\delta + \epsilon - \zeta_i - \zeta_j|^+)^2}{2(n_i\hat{\sigma}^{2, \alpha}_i + n_j\hat{\sigma}^{2, \alpha}_j)}\Big) + \alpha 
\end{equation}
where $n_i$ and $n_j$ are the number of sampled trajectories in $\mathcal{C}_i$ and $\mathcal{C}_j$, respectively. 
An upper bound on the return variance is given by $\hat{\sigma}^{2, \alpha}$ with significance $\alpha \in (0, 1)$.
The $\zeta$ terms are upper bounds on the expected bias of the baseline value estimate.
\end{corollary}

Intuitively, the action error provides a limit on how likely it is that an action from the search set is better than the action with the maximum estimated value.
Though these bounds decay exponentially with the number of samples, empirical experiments show they are fairly loose. 
We therefore present an additional set of bounds and error approximations that hold more tightly for appropriate problems. 

\subsection{Central Limit Theorem Bound}
The previously presented inequalities only assumed the values estimates were bounded. 
Though widely applicable, the presented bounds were found to be fairly loose. 
We develop tighter bounds by assuming the distribution of return estimates follows a normal distribution for cases in which the central limit theorem (CLT) applies.
The central limit theorem is valid for estimates over stationary distributions, but not necessarily for non-stationary process.
The CLT has been shown to hold for payouts from non-stationary Markov chains for a wide range of cases.
By applying the CLT with a finite sample penalty, we can derive tighter versions of the bounds in~\cref{thm: gen as bound} and~\cref{thm: gen ae bound}. 

\begin{theorem}~\label{thm: clt as bound}
For a Monte Carlo search satisfying the central limit theorem returning action value estimate $\bar{Q}_i$, the probability that the true optimal action value is overestimated $\bar{Q}_i - Q^*_i \geq \epsilon$ is no greater than
\begin{equation}~\label{eq: clt se bound}
1 - \Phi(\epsilon - \zeta_i; 0, \hat{\sigma}_i^{2, \alpha}/n_i) + \alpha +  \frac{0.4748}{n_i}
\end{equation}
where $\Phi(x; 0, \sigma^2)$ gives the cumulative distribution function for a zero-mean Gaussian with variance $\sigma^2$. 
\end{theorem}
This bound is built on the Gaussian assumption of the CLT with two penalties for finite samples. 
The first is the variance upper bound previously presented in~\cref{eq: var bound}.
The second penalty is introduce in the far right term. 
This term is a correction for CLT convergence based on the Berry-Esseen theorem~\citep{berry1941, esseen1942} with the constant term proved by~\citet{shevtsova2011}.

As before, we present the corollary bound on error probability. 
\begin{corollary}~\label{thm: clt ae bound}
Let a Monte Carlo solver generate estimates such that the central limit theorem holds. 
For a search that returns action value estimates over two actions $\bar{Q}_i - \bar{Q}_j = \delta$, the probability that the true action value gap $Q_i - Q_j \leq -\epsilon$ is no greater than
\begin{equation}~\label{eq: clt ae bound}
1 - \Phi(\delta + \epsilon - \zeta_i - \zeta_j; 0, \bar{\sigma}^{2, \alpha}) + \alpha +  \frac{0.4748}{n_i + n_j}
\end{equation}
where the variance upper bound is given by $\bar{\sigma}^{2, \alpha} = \hat{\sigma}^{2,\alpha}_i/n_i + \hat{\sigma}^{2,\alpha}_j/n_j$.
\end{corollary}

\subsection{Error Estimation}
Instead of computing an upper bound, which may be loose, it may be beneficial to instead compute an estimate of the error probability. 
An error estimate that holds to the true value more tightly than a bound may be more useful for certain applications, such as for a stopping criteria during anytime search.
We can model the distribution of mean estimates of samples drawn from an unknown distribution using Student's t-distribution~\citep{student1908}.
To approximate the value error, a similar estimate is defined by 
\begin{equation}~\label{eq: t subopt est}
1 - T_{n-1}\Big(\frac{\epsilon - \zeta_i}{(\sqrt{V_i}/n_i)}\Big)
\end{equation}
which follows from the single-variable t-distribution with $n-1$ degrees of freedom.

We can similarly estimate the action error as
\begin{equation}~\label{eq: t ae est}
1 - T_{\nu}\Big(\frac{\delta + \epsilon - \zeta_i - \zeta_j}{(\sqrt{V_i}/n_i + \sqrt{V_j}/n_j)}\Big)
\end{equation}
where $T_{\nu}(x)$ is the cumulative distribution function for the t-distribution with $\nu$ degrees of freedom. 
For estimates over the difference of two variables, we approximate the degrees of freedom as
\begin{equation}
    \nu = \frac{\big(V_i/n_i + V_j/n_j\big)^2}{V_i^2/n_i^2 + V_j/n_j^2}
\end{equation}
following the method proposed by~\citet{welch1947}.

\section{Numerical Evaluation}
Experiments were conducted to evaluate the tightness of the presented bounds and the accuracy of the error approximation. 
We tested two problems: a simple multi-armed bandit and a grid-world MDP.
We calculated the bounds under various conditions using a simple Monte Carlo estimator for both problems. 
For the MDP, we also tested MCTS with upper confidence tree (UCT) sampling. 
All testing was done in the Julia language using the POMDPs.jl framework~\citep{Egorov2017}.
The $\alpha$ were minimized with Nelder-Meade optimization~\citep{nelder1965}.

The simple Monte Carlo solver draws $n$ samples of actions from a discrete distribution over the complete action space.
Distribution weights for each action $w_i$ are calculated from the softmax over the value estimates as
\begin{equation}
    w_i \gets \frac{\exp{\bar{Q}_i \tau^{-1}}}{\sum_i \exp{\bar{Q}_i \tau^{-1}}}
\end{equation}
where $\tau$ is a temperature parameter, which was set to 10 for experiments in this work. 
Actions that have not yet been sampled have $q_i \gets \infty$.
In the MDP setting, after sampling the first action, the remaining actions are selected according to an $\epsilon$-greedy baseline policy, up to a specified depth $h$.
Source code for the experiments is provided in the supplementary materials. 

\subsection{Multi-Armed Bandit}
The bandit problem had 10 arms, each with a Gaussian payout distribution.
The tests were run in episodes, with each episode having differently parameterized distributions.
The payout mean values were sampled from a uniform distribution on $[-1, 1]$.
Each payout distribution had the same variance $\sigma^2 = 0.5$ each episode. 
The bandit expected payouts were estimated with the Monte Carlo solver for varying numbers of samples $n$.

\begin{figure*}[ht]
     \centering
     \subfigure[]{\includegraphics[width=0.49\textwidth]{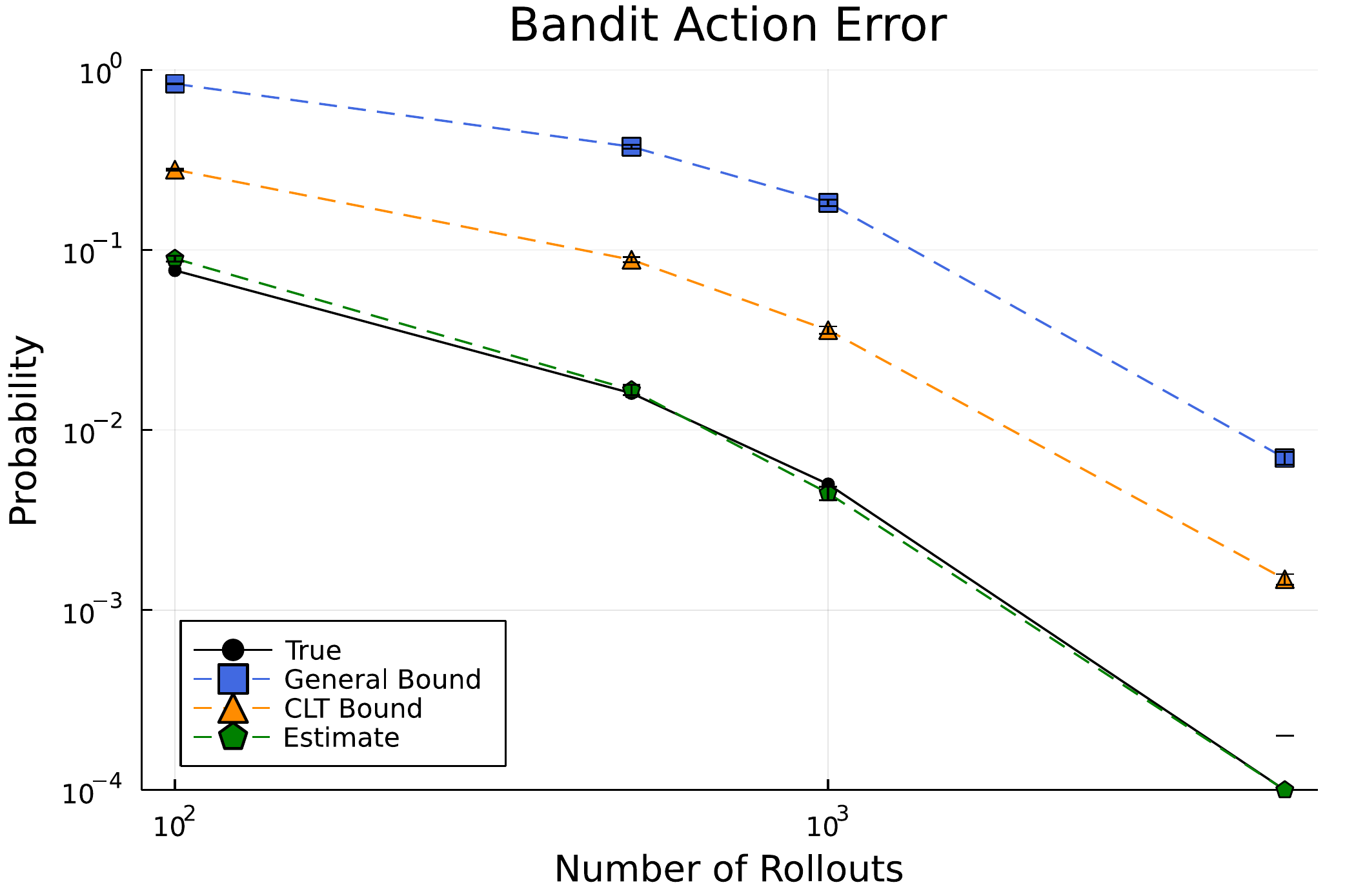}}\label{fig: bandit err 0.5}
     \hfill
     \subfigure[]{\includegraphics[width=0.49\textwidth]{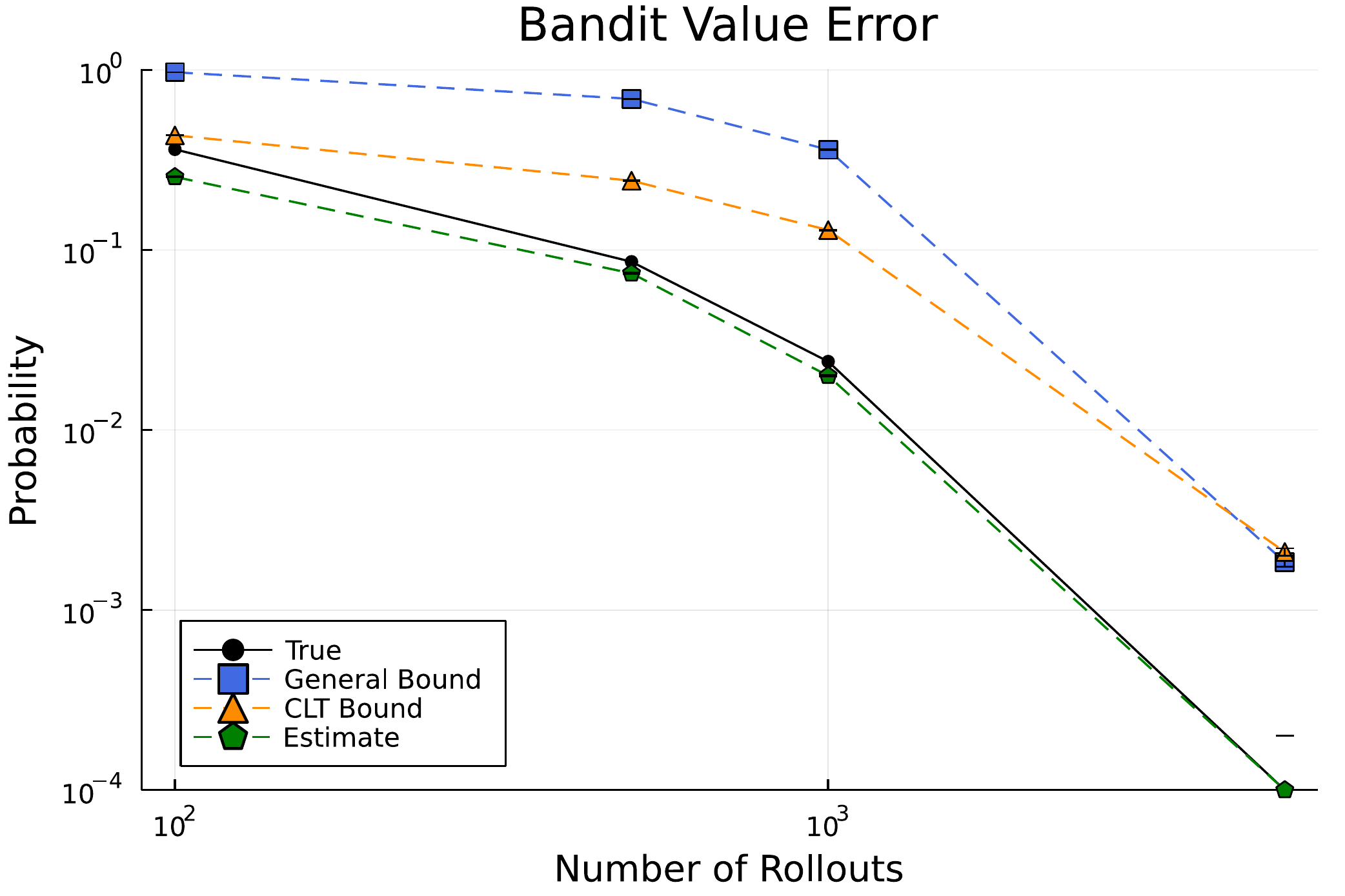}}\label{fig: bandit sub 0.5}
     \caption{Bandit experiment results. Each graph gives the calculated general upper bound, the CLT bound, the error rate estimation, and the true observed error rate for varying numbers of sample rollouts. Figure (a) gives the action error and figure (b) gives the value error.}
     \label{fig: bandit results}
\end{figure*}

For each episode, the error probability bound was calculated for the two actions with the highest estimated values.
The sub-optimality bound was calculated for the best action for each sampling level and variance.
The margin was $\epsilon = 0.1$ for both experiments.
We calculated the true rates of error and sub-optimality in excess of $\epsilon$ across episodes. 
The average upper bound calculated for each trial and one standard error bounds are plotted along with the true rate curves in~\cref{fig: bandit results}.
The x-axis plots the total number of samples for the the search, not necessarily the samples over the measured action.

As can be seen, both the error and sub-optimality bounds decay exponentially with the number of samples drawn.
The CLT bounds are tighter than the general bounds in both cases for all sampled points. 
The error estimate is significantly closer than either bound, though it does under-estimate the observed rate a several points.

\subsection{Grid World MDP}
In the grid world MDP, an agent is required to navigate through a discrete, 2D world from a random initial position to a goal destination. 
There are two goal locations generating positive rewards of 10 and 3, and two penalty locations giving negative rewards of -10 and -5.
The episode terminates when the agent reaches any of the four reward-generating locations.
Each step, the agent takes an action to attempt to move in one of the four cardinal directions.
The agent successfully moves in the intended direction with a probability of 0.7 and moves randomly otherwise.
The grid tested was $10 \times 10$.
The time discount was set to $\gamma = 0.95$.

\subsubsection{Simple MC Solver}
\begin{figure*}[h]
     \centering
     \subfigure[]{\includegraphics[width=0.49\textwidth]{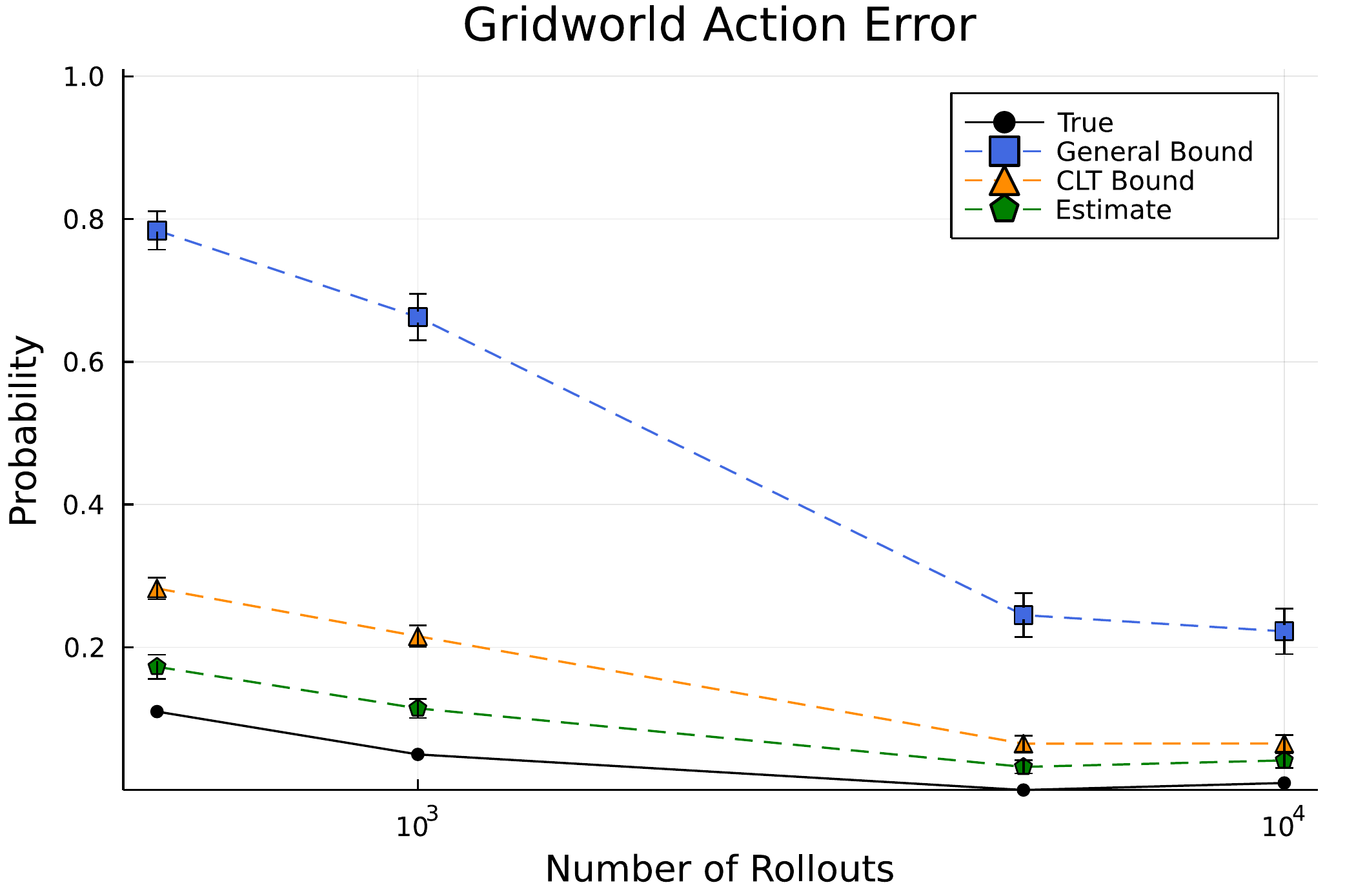}}\label{fig: grid_td err 10}
     \hfill
     \subfigure[]{\includegraphics[width=0.49\textwidth]{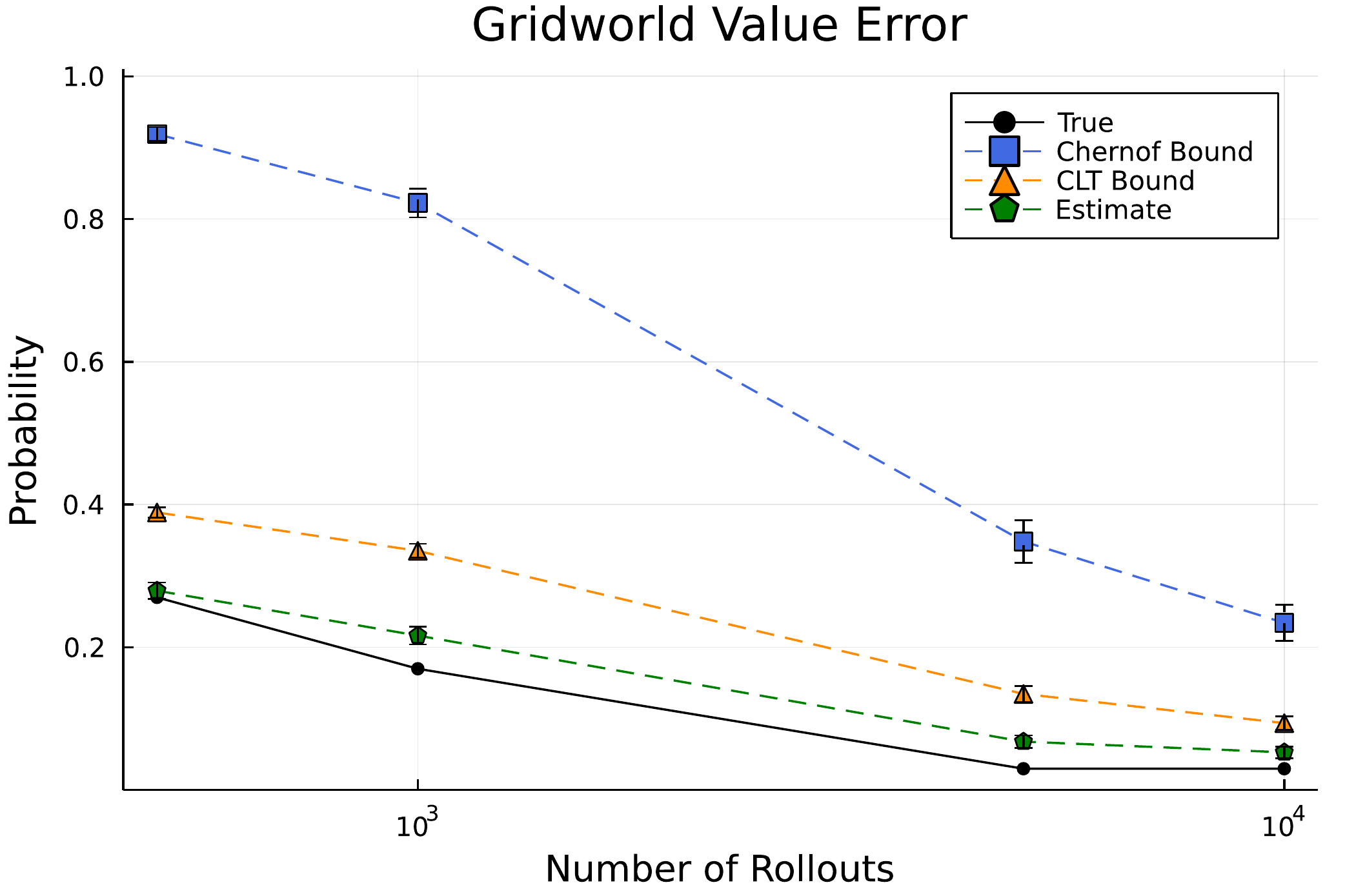}}\label{fig: grid_td sub 10}
     \caption{Grid world MC solver results. Each graph gives the calculated general Chernoff upper bound, the CLT bound, the error rate estimation, and the true observed error rate for varying numbers of sample rollouts. Figure (a) gives the action error and figure (b) gives the value error. }
     \label{fig: gridworld td results}
\end{figure*}

We used offline value iteration to learn a baseline policy and approximate optimal action values. 
The offline solver terminated after an update step with a Bellman residual less than $10^{-6}$.
The simple Monte Carlo solver used an $\epsilon$-greedy baseline policy select actions after the initial step. 
The bootstrap value estimates were generated by the baseline policy with additive uniform noise bounded between $-0.1$ and $0.1$. 
We ran the tests with the solver searching to depths of 5, 10, and 25. 
As with the bandit problem, we calculated the error and sub-optimality rates for the top performing actions. 
The results for depth 10 search are shown in~\cref{fig: gridworld td results}.

As with the bandit results, both the error and sub-optimality bounds decay exponentially with the number of samples drawn, and the CLT bounds hold more tightly.
At low samples, the bounds are tighter for the MDP than for the bandit problem, due to the higher true error rates.
Unlike the bandit problem, the t-distribution estimate does not underestimate the observed error rate at any point. 
This is likely due to the fact that the t-distribution estimate does still introduce a conservative error assumption in the bounds used to calculate the leaf node bias $\zeta$ terms.
Using an estimate of leaf node error instead of an upper bound could result in more accurate estimates, though with more frequent under-approximation.

\subsubsection{Monte Carlo Tree Search}
\begin{figure*}[h]
     \centering
     \subfigure[]{\includegraphics[width=0.49\textwidth]{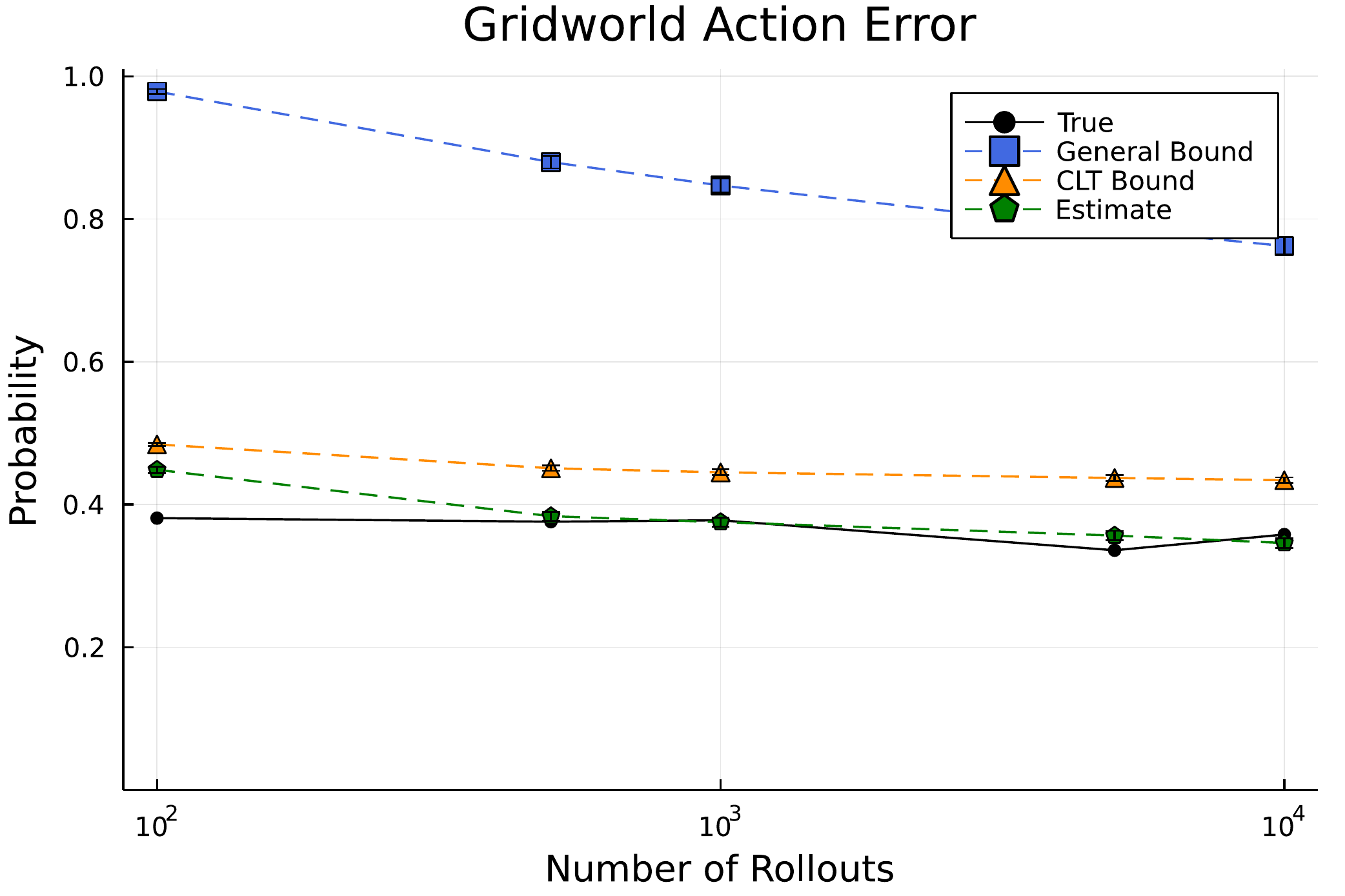}}\label{fig: grid_mcts err}
     \hfill
     \subfigure[]{\includegraphics[width=0.49\textwidth]{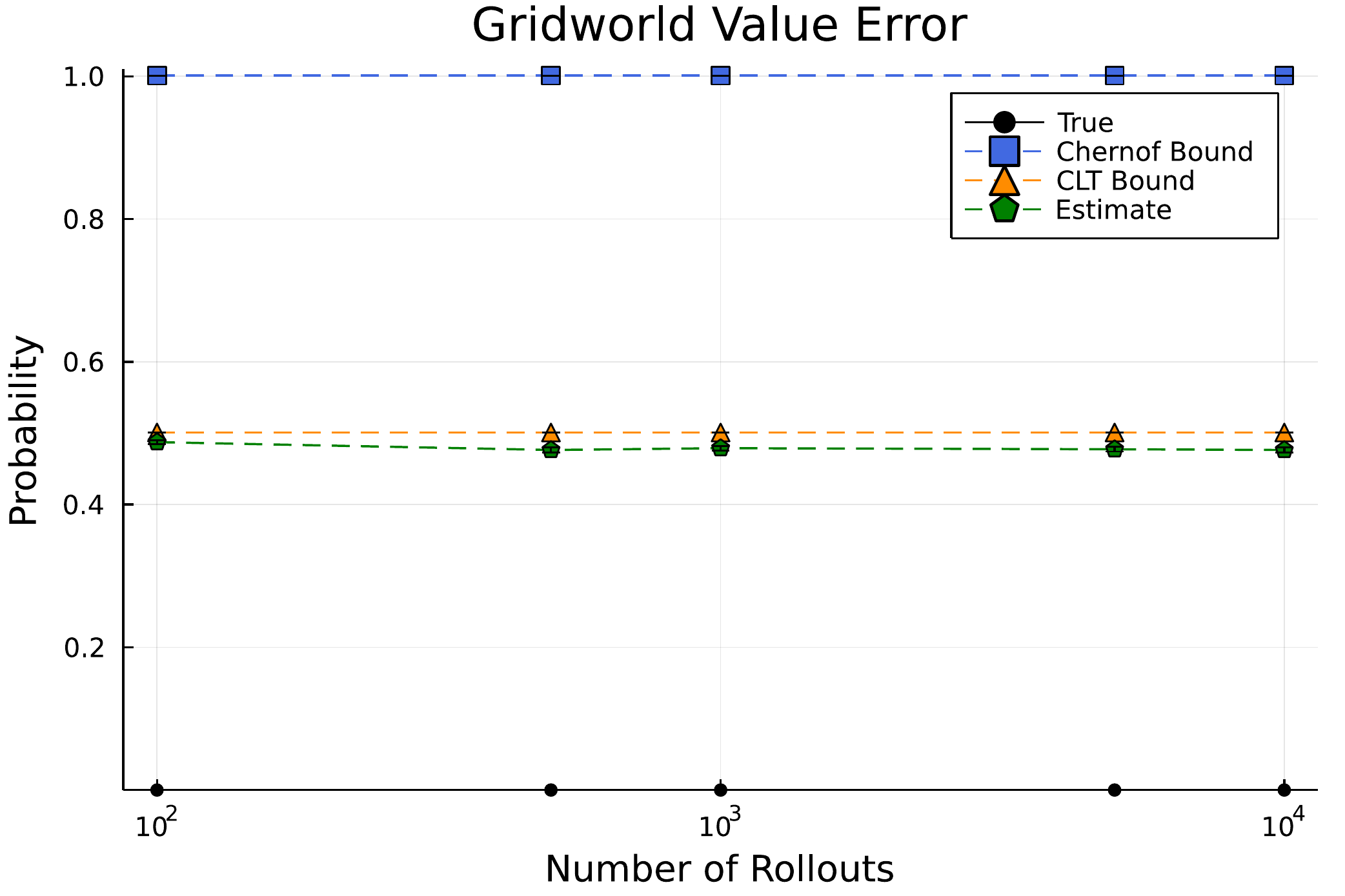}}\label{fig: grid_mcts sub}
     \caption{Gridworld MCTS solver results. Each graph gives the calculated general upper bound, the CLT bound, the error rate estimation, and the true observed error rate for varying numbers of sample rollouts. Figure (a) gives the action error and figure (b) gives the value error.}
     \label{fig: gridworld mcts results}
\end{figure*}

We tested the tightness of the bounds on the popular MCTS-UCT algorithm, using the MCTS.jl implementation.
We tested the performance with max depth set to 25 steps for each search.
The UCT exploration constant was set to 1.0.
As with the simple solver, leaf node values were estimated by the baseline policy with additive uniform noise.
The results are shown in~\cref{fig: gridworld mcts results}.

The same trends can be observed in the MCTS results as in the simple Monte Carlo search results, however, the bounds are significantly looser at high sample counts in the MCTS tests.
This is likely due to the influence of several shallow searches early in the search leading to higher magnitude $\zeta$ terms. 
Unlike the simple solver, the UCT algorithm does not satisfy the convergence condition, as value estimates include results from early search results during exploration. 
This leads to a well known underestimation problem, often corrected through post-search max-value backup. 
This leads to the observed rate of over-estimation being zero throughout search. 
Including early samples in the estimates also means that the samples are not from a stationary distribution, and that the CLT may not necessarily apply. 
Despite this, the CLT bound does not underestimate the true value at any search setting. 

These results suggest several adaptations to MCTS methods to improve expected error bounds.
For a given number of samples, tighter error bounds can be achieved when the sample counts of the compare actions are comparable.
For situations where accurate action selection from a discrete set is important, this suggests that increased exploration at the end of search may improve error performance. 
Deeper searches tend to result in better estimates than shallow searches. 
This suggests modifications to either increase search depth or to more heavily weight samples from longer trajectories in the value estimate. 

\section{Conclusions and Future Work}
In this work, we presented empirical bounds on the error probability for convergent Monte Carlo solvers on non-stationary bandit problems and Markov decision processes.
We presented bounds for general, bounded-value problems with convergent solver and tighter bounds applicable when the central limit theorem holds. 
To provide a useful metric for anytime search, we also proposed a simple error approximation based on the t-distribution.
We demonstrated that these bounds hold for a multi-armed bandit and a discrete grid world MDP for both a simple Monte Carlo estimator and a more complex tree search.
Results showed that the bounds are loose for low sample numbers, but decay exponentially as samples increase. 
The results also show that the CLT bounds tend to be tighter than the general bounds, and that the error estimate is a good approximation of the true rate in the studied problems.
We anticipate that these measures may be especially useful to methods that use an informed baseline such as offline policy improvement and performance verification of safety-critical systems.

The current work only tested these bounds for simple bandits and MDPs. 
Future work shall explore their performance in other, more complex tasks such as MCTS for partially observable MDPs~\citep{silver2010}.
Even with a moderate number of samples, the bounds were sometimes loose. 
Future work will look to specialize the general form presented in this work to produce tighter bounds for specific algorithms. 
The current work also only considered arithmetic mean estimators. 
Future work will extend the bounds to more general estimators. 

\bibliography{bib}

\begin{thebibliography}{33}
\providecommand{\natexlab}[1]{#1}
\providecommand{\url}[1]{\texttt{#1}}
\expandafter\ifx\csname urlstyle\endcsname\relax
  \providecommand{\doi}[1]{doi: #1}\else
  \providecommand{\doi}{doi: \begingroup \urlstyle{rm}\Url}\fi

\bibitem[Aras et~al.(2019)Aras, Lee, Pananjady, and Courtade]{aras2019}
Efe Aras, Kuan{-}Yun Lee, Ashwin Pananjady, and Thomas~A. Courtade.
\newblock A family of {B}ayesian {C}ram{\'{e}}r-{R}ao bounds, and consequences
  for log-concave priors.
\newblock In \emph{{IEEE} International Symposium on Information Theory
  ({ISIT})}, pages 2699--2703, 2019.
\newblock \doi{10.1109/ISIT.2019.8849630}.

\bibitem[Arlotto and Steele(2016)]{arlotto2016}
Alessandro Arlotto and J.~Michael Steele.
\newblock A central limit theorem for temporally nonhomogenous {M}arkov chains
  with applications to dynamic programming.
\newblock \emph{Mathematics of Operations Research}, 41\penalty0 (4):\penalty0
  1448--1468, 2016.

\bibitem[Audibert et~al.(2009)Audibert, Munos, and
  Szepesv{\'{a}}ri]{audibert2009}
Jean{-}Yves Audibert, R{\'{e}}mi Munos, and Csaba Szepesv{\'{a}}ri.
\newblock Exploration-exploitation tradeoff using variance estimates in
  multi-armed bandits.
\newblock \emph{Theoretical Computational Science}, 410\penalty0 (19):\penalty0
  1876--1902, 2009.
\newblock \doi{10.1016/j.tcs.2009.01.016}.

\bibitem[Auer et~al.(2002)Auer, Cesa{-}Bianchi, and Fischer]{auer2002}
Peter Auer, Nicol{\`{o}} Cesa{-}Bianchi, and Paul Fischer.
\newblock Finite-time analysis of the multiarmed bandit problem.
\newblock \emph{Journal of Machine Learning Research}, 47\penalty0
  (2--3):\penalty0 235--256, 2002.

\bibitem[Bernstein(1927)]{Bernstein1927}
Serge Bernstein.
\newblock Sur l'extension du théoréme limite du calcul des probabilités aux
  sommes de quantités dépendantes.
\newblock \emph{Mathematische Annalen}, 97\penalty0 (1):\penalty0 1432--1807,
  1927.
\newblock \doi{10.1007/BF01447859}.

\bibitem[Berry(1941)]{berry1941}
Andrew~C Berry.
\newblock The accuracy of the {G}aussian approximation to the sum of
  independent variates.
\newblock \emph{Transactions of the American Mathematical Society}, 49\penalty0
  (1):\penalty0 122--136, 1941.

\bibitem[Beygelzimer et~al.(2011)Beygelzimer, Langford, Li, Reyzin, and
  Schapire]{beygelzimer2011}
Alina Beygelzimer, John Langford, Lihong Li, Lev Reyzin, and Robert~E.
  Schapire.
\newblock Contextual bandit algorithms with supervised learning guarantees.
\newblock In \emph{International Conference on Artificial Intelligence and
  Statistics ({AISTATS})}, volume~15, pages 19--26, 2011.

\bibitem[Dobrushin(1956)]{dobrushin1956}
Roland~L Dobrushin.
\newblock Central limit theorem for nonstationary {M}arkov chains.
\newblock \emph{Theory of Probability \& Its Applications}, 1\penalty0
  (1):\penalty0 65--80, 1956.

\bibitem[Egorov et~al.(2017)Egorov, Sunberg, Balaban, Wheeler, Gupta, and
  Kochenderfer]{Egorov2017}
Maxim Egorov, Zachary~N. Sunberg, Edward Balaban, Tim~Allan Wheeler, Jayesh~K.
  Gupta, and Mykel~J. Kochenderfer.
\newblock {POMDP}s.jl: {A} framework for sequential decision making under
  uncertainty.
\newblock \emph{Journal of Machine Learning Research}, 18:\penalty0 26:1--26:5,
  2017.

\bibitem[Esseen(1942)]{esseen1942}
Carl-Gustav Esseen.
\newblock On the {L}iapunoff limit of error in the theory of probability.
\newblock \emph{Arkiv for matematik, astronomi och fysik, A: 1--19}, 1942.

\bibitem[Feldman and Domshlak(2014)]{feldman2014}
Zohar Feldman and Carmel Domshlak.
\newblock Simple regret optimization in online planning for {M}arkov decision
  processes.
\newblock \emph{Journal of Artificial Intelligence Research}, 51:\penalty0
  165--205, 2014.
\newblock \doi{10.1613/jair.4432}.

\bibitem[Gosset(1908)]{student1908}
William Sealy~(Student) Gosset.
\newblock The probable error of a mean.
\newblock \emph{Biometrika}, 6\penalty0 (1):\penalty0 1--25, 1908.
\newblock ISSN 00063444.

\bibitem[Grill et~al.(2016)Grill, Valko, and Munos]{grill2016}
Jean{-}Bastien Grill, Michal Valko, and R{\'{e}}mi Munos.
\newblock Blazing the trails before beating the path: Sample-efficient
  {M}onte-{C}arlo planning.
\newblock In \emph{Advances in Neural Information Processing Systems
  (NeurIPS)}, pages 4673--4681, 2016.

\bibitem[Grill et~al.(2019)Grill, Domingues, M{\'{e}}nard, Munos, and
  Valko]{grill2019}
Jean{-}Bastien Grill, Omar~Darwiche Domingues, Pierre M{\'{e}}nard, R{\'{e}}mi
  Munos, and Michal Valko.
\newblock Planning in entropy-regularized {M}arkov decision processes and
  games.
\newblock In \emph{Advances in Neural Information Processing Systems
  (NeurIPS)}, pages 12383--12392, 2019.

\bibitem[{H\"older}(1889)]{holder1889}
Otto {H\"older}.
\newblock {Ueber einen Mittelwertsatz}.
\newblock \emph{{G\"ott. Nachr.}}, 1889:\penalty0 38--47, 1889.

\bibitem[Kaufmann and Koolen(2017)]{kaufmann2017}
Emilie Kaufmann and Wouter~M. Koolen.
\newblock Monte-{C}arlo tree search by best arm identification.
\newblock In \emph{Advances in Neural Information Processing Systems
  (NeurIPS)}, pages 4897--4906, 2017.

\bibitem[Kearns et~al.(1999)Kearns, Mansour, and Ng]{kearns1999}
Michael~J. Kearns, Yishay Mansour, and Andrew~Y. Ng.
\newblock A sparse sampling algorithm for near-optimal planning in large
  {M}sarkov decision processes.
\newblock In \emph{International Joint Conference on Artificial Intelligence
  (IJCAI)}, pages 1324--1231, 1999.

\bibitem[Kocsis and Szepesv{\'{a}}ri(2006)]{kocsis2006}
Levente Kocsis and Csaba Szepesv{\'{a}}ri.
\newblock Bandit based {M}onte-{C}arlo planning.
\newblock In \emph{European Conference on Machine Learning (ECML)}, 2006.

\bibitem[Lai and Robbins(1985)]{lai1985}
Tze~Leung Lai and Herbert Robbins.
\newblock Asymptotically efficient adaptive allocation rules.
\newblock \emph{Advances in Applied Mathematics}, 6\penalty0 (1):\penalty0
  4--22, 1985.

\bibitem[Markov(1951)]{markov1951}
Andrey~A. Markov.
\newblock Investigations of general experiments connected by a {M}arkov chain.
\newblock \emph{Collected Works}, pages 465--509, 1951.

\bibitem[Maurer and Pontil(2009)]{maurer2009}
Andreas Maurer and Massimiliano Pontil.
\newblock Empirical {B}ernstein bounds and sample-variance penalization.
\newblock In \emph{Conference on Learning Theory ({COLT})}, 2009.

\bibitem[Nelder and Mead(1965)]{nelder1965}
John~A. Nelder and R.~Mead.
\newblock A simplex method for function minimization.
\newblock \emph{Computer Journal}, 7\penalty0 (4):\penalty0 308--313, 1965.
\newblock \doi{10.1093/comjnl/7.4.308}.

\bibitem[Peel et~al.(2013)Peel, Anthoine, and Ralaivola]{peel2013}
Thomas Peel, Sandrine Anthoine, and Liva Ralaivola.
\newblock Empirical {B}ernstein inequality for martingales : Application to
  online learning.
\newblock \emph{HAL}, 2013.

\bibitem[Shah et~al.(2020)Shah, Xie, and Xu]{shah2020}
Devavrat Shah, Qiaomin Xie, and Zhi Xu.
\newblock Non-asymptotic analysis of {M}onte {C}arlo tree search.
\newblock In \emph{Joint International Conference on Measurement and Modeling
  of Computer Systems ({SIGMETRICS})}, pages 31--32, 2020.
\newblock \doi{10.1145/3393691.3394202}.

\bibitem[Shevtsova(2011)]{shevtsova2011}
I.~Shevtsova.
\newblock On the absolute constants in the {B}erry-{E}sseen-type inequalities.
\newblock \emph{Doklady Mathematics}, 89:\penalty0 378--381, 2011.

\bibitem[Silver and Veness(2010)]{silver2010}
David Silver and Joel Veness.
\newblock Monte-{C}arlo planning in large {POMDP}s.
\newblock In \emph{Advances in Neural Information Processing Systems (NIPS)},
  2010.

\bibitem[Silver et~al.(2016)Silver, Huang, Maddison, Guez, Sifre, van~den
  Driessche, Schrittwieser, Antonoglou, Panneershelvam, Lanctot, Dieleman,
  Grewe, Nham, Kalchbrenner, Sutskever, Lillicrap, Leach, Kavukcuoglu, Graepel,
  and Hassabis]{silver2016}
David Silver, Aja Huang, Chris~J. Maddison, Arthur Guez, Laurent Sifre, George
  van~den Driessche, Julian Schrittwieser, Ioannis Antonoglou, Vedavyas
  Panneershelvam, Marc Lanctot, Sander Dieleman, Dominik Grewe, John Nham, Nal
  Kalchbrenner, Ilya Sutskever, Timothy~P. Lillicrap, Madeleine Leach, Koray
  Kavukcuoglu, Thore Graepel, and Demis Hassabis.
\newblock Mastering the game of {G}o with deep neural networks and tree search.
\newblock \emph{Nature}, 529\penalty0 (7587):\penalty0 484--489, 2016.
\newblock \doi{10.1038/nature16961}.

\bibitem[Silver et~al.(2017)Silver, Schrittwieser, Simonyan, Antonoglou, Huang,
  Guez, Hubert, Baker, Lai, Bolton, Chen, Lillicrap, Hui, Sifre, van~den
  Driessche, Graepel, and Hassabis]{silver2017}
David Silver, Julian Schrittwieser, Karen Simonyan, Ioannis Antonoglou, Aja
  Huang, Arthur Guez, Thomas Hubert, Lucas Baker, Matthew Lai, Adrian Bolton,
  Yutian Chen, Timothy~P. Lillicrap, Fan Hui, Laurent Sifre, George van~den
  Driessche, Thore Graepel, and Demis Hassabis.
\newblock Mastering the game of go without human knowledge.
\newblock \emph{Nature}, 550\penalty0 (7676):\penalty0 354--359, 2017.
\newblock \doi{10.1038/nature24270}.

\bibitem[Sinn and Chen(2013)]{sinn2013}
Mathieu Sinn and Bei Chen.
\newblock Central limit theorems for conditional {M}arkov chains.
\newblock In \emph{International Conference on Artificial Intelligence and
  Statistics (AISTATS)}, pages 554--562, 2013.

\bibitem[Sutton and Barto(2018)]{sutton1998}
Richard~S. Sutton and Andrew~G. Barto.
\newblock \emph{Reinforcement Learning: An Introduction}.
\newblock MIT Press, 2 edition, 2018.

\bibitem[Tichavsky et~al.(1998)Tichavsky, Muravchik, and
  Nehorai]{tichavksy1998}
P.~Tichavsky, C.H. Muravchik, and A.~Nehorai.
\newblock Posterior {C}ram\'er-{R}ao bounds for discrete-time nonlinear
  filtering.
\newblock \emph{IEEE Transactions on Signal Processing}, 46\penalty0
  (5):\penalty0 1386--1396, 1998.
\newblock \doi{10.1109/78.668800}.

\bibitem[Welch(1947)]{welch1947}
Bernard~Lewis Welch.
\newblock {The Generalization of {S}tudent's problem when several different
  population variances are involved}.
\newblock \emph{Biometrika}, 34\penalty0 (1-2):\penalty0 28--35, 1947.
\newblock ISSN 0006-3444.
\newblock \doi{10.1093/biomet/34.1-2.28}.

\bibitem[Zhang et~al.(2021)Zhang, Yang, Ji, and Du]{zhang2021}
Zihan Zhang, Jiaqi Yang, Xiangyang Ji, and Simon~S. Du.
\newblock Variance-aware confidence set: Variance-dependent bound for linear
  bandits and horizon-free bound for linear mixture {MDP}.
\newblock \emph{Computing Research Repository}, 2021.

\end{thebibliography}

\appendix

\section{Proofs}
This section provides proofs the presented action error bounds and shows how the same procedure can be applied to prove the value error bound.
We first present several lemmas.
The first lemma presents a bound on the difference in expected error of action value estimates generated by a Monte Carlo search. 
\begin{lemma}~\label[lemma]{lem: exp error}
    Given a Monte Carlo estimate $\bar{Q}_i$ of true value $Q_i$, the expected error is 
    \begin{equation}
        \mathbb{E}[\bar{Q}_i - Q_i] \leq \frac{1}{|\mathcal{C}_i|}\sum_k \gamma^{H_k}\epsilon^0_k 
    \end{equation}
    where $\epsilon_k^0$ is the expected error of the baseline estimate at the end of the trajectory $k$ of depth $H_k$.
\end{lemma}
A proof of~\cref{lem: exp error} is given in~\cref{sec: lem proof}. 
The next lemma shows that an appropriate upper bound on the conditional likelihood of an estimator of an unknown parameter is also an upper bound on a probability of the parameter. 
\begin{lemma}~\label[lemma]{lem: cramer}
    Let $\theta$ be a population parameter and let $\bar{\theta}$ be a potentially biased estimator of that parameter with variance $\bar{\sigma}^2$. 
    If the probability 
    \begin{align}
        P(\bar{\theta} \geq a \mid  \theta = b) \leq c(\bar{\sigma}^2)
    \end{align}
    holds, where $c(\bar{\sigma}^2)$ is a function that increases monotonically with $\bar{\sigma}^2$, then the inequality
    \begin{align}
        P(\theta \geq b \mid \bar{\theta} = a) \leq c(\bar{\sigma}^2)
    \end{align}
    also holds.
\end{lemma}
\Cref{lem: cramer} follows from the Bayesian Cram\'er-Rao inequality~\citep{tichavksy1998, aras2019}, which, for appropriate priors, implies
\begin{equation}
    \mathrm{Var}[\theta \mid X] \leq \mathrm{Var}[\bar{\theta} \mid X]
\end{equation}
for distributional parameter $\theta$, estimator $\bar{\theta}$, and data $X$.
The next lemma bounds the true variance of an expectation estimator around the sample variance. 
\begin{lemma}~\label[lemma]{lem: variance}
    Let $\bar{X} = \frac{1}{n}\sum_i^n X_i$ be an estimator of the mean over $n$ random variables with variance $\bar{\sigma}^2$.
    Given a estimate $\bar{x} = \frac{1}{n}\sum_i^n x_i \sim X_i$ and sample variance $V_n = \frac{1}{n}\sum_i^n (x_i - \bar{x})^2$, the variance is bounded as
    \begin{equation}
        \bar{\sigma}^2 \leq \frac{b^2}{n}\Bigg(\frac{1}{b}V_n^\frac{1}{2} + \sqrt{\frac{-\ln{\alpha}}{n - 1}}  \Bigg)^2
    \end{equation}
    where $b = \max_i x_i - \min_i x_i$ is the range of the sample values and $\alpha \in (0, 1)$ is a significance level.
\end{lemma}
\Cref{lem: variance} follows from the sample variance penalty for empirical Bernstein bounds stated in theorem 10 of the work by~\citet{maurer2009}.
The final lemma provides an upper bound to the finite sample penalty of the Berry-Esseen theorem when the true second and third moments are not known.
\begin{lemma}~\label[lemma]{lem: berry-esseen}
    Given a set of zero-mean $i.i.d.$ random variables $X_1, \dots, X_n$ with variance $\sigma^2$ and $Y = 1/n sum_i X_i$, the difference in between the true cumulative distribution function $F_n$ and the normal Gaussian CDF $\Phi$ is bounded as
    \begin{equation}
        \sup_{x \in \mathbb{R}}|F_n(x) - \Phi(x)| \leq \frac{C}{n} 
    \end{equation}
    where for all $n > 0$ and some value of $C \leq 0.4748$.
\end{lemma}

\subsection{Proof of~\Cref{thm: gen as bound}}
\begin{proof}~\label{proof: gen as}
    The proof begins by constructing defining a bound as
\begin{align}
    & P\big(\bar{Q}_i - Q_i \geq \epsilon \big)~\label{proof start 1} \\
    = & P\big(\bar{Q}_i - Q_i - \mathbb{E}\big[\bar{Q}_i \big] \geq \epsilon - \mathbb{E}\big[\bar{Q}_i \big] \big) \\
    = & P\big(\bar{Q}_i - \mathbb{E}\big[\bar{Q}_i \big] \geq \epsilon - \mathbb{E}\big[\bar{Q}_i - Q_i\big] \big) \\
    \leq & P\big(\bar{Q}_i - \mathbb{E}\big[\bar{Q}_i \big] \geq \epsilon - \sum_{k \in Ch_i}\gamma^{H_k}\epsilon^0_k \big)~\label{error step 1} \\
    \leq & P\big(\bar{Q}_i - \mathbb{E}\big[\bar{Q}_i \big] \geq \epsilon - \sum_{k \in Ch_i}\gamma^{H_k}\epsilon^U_k \big)~\label{error bound step 1} 
\end{align}
where~\cref{error step 1} follows from~\cref{lem: exp error}.
The expected baseline estimator error $\epsilon^0$ is not generally known, though its magnitude can be limited by a term $\epsilon^U \geq |\epsilon^0|$.
Applying this inequality results in~\cref{error bound step 1}.
We can bound the probability of the zero-mean quantity in~\cref{error bound step 1} with a Chernoff-Hoeffding bound as
\begin{align}
    P\big(\bar{Q}_i - Q_i \geq \epsilon \big)
    \leq & \exp{\Big(\frac{-2n_i (|\epsilon - \zeta|^+)^2}{n_i\bar{\sigma}_i^2}\Big)}~\label{chernoff step 1}
\end{align}
where $\bar{\sigma}^2$ gives the variance of the value estimators.
Since the true variance is not generally known, we apply~\cref{lem: variance} with a union bound to arrive at
\begin{align}
    \leq & \exp{\Big(\frac{-2n_i (|\epsilon - \zeta|^+)^2}{n_i\hat{\sigma}_i^2}\Big)} + \alpha~\label{variance step}
\end{align}
which concludes the proof.
\end{proof}

\subsection{Proof of~\Cref{thm: gen ae bound}}
\begin{proof}~\label{proof: gen ae}
    The proof begins by constructing a bound on the likelihood of search results as
\begin{align}
    & P\big(\bar{Q}_i - \bar{Q}_j \geq \delta \mid Q^\pi_j - Q^\pi_i = \epsilon \big)~\label{proof start} \\
    = & P\big(\bar{Q}_i - \bar{Q}_j - \mathbb{E}\big[\bar{Q}_i - \bar{Q}_j \big] \geq \delta - \mathbb{E}\big[\bar{Q}_i - \bar{Q}_j \big] \big) \\
    = & P\big(\bar{Q}_i - \bar{Q}_j - \mathbb{E}\big[\bar{Q}_i - \bar{Q}_j \big] \geq \delta + \epsilon - \mathbb{E}\big[\bar{Q}_i - Q_i\big] + \mathbb{E}\big[\bar{Q}_j - Q_j\big] \big) \\
    \leq & P\big(\bar{Q}_i - \bar{Q}_j - \mathbb{E}\big[\bar{Q}_i - \bar{Q}_j \big] \geq \delta + \epsilon - \sum_{k \in Ch_i}\gamma^{H_k}\epsilon^0_k + \sum_{l \in Ch_j}\gamma^{H_l}\epsilon^0_l \big)~\label{error step} \\
    \leq & P\big(\bar{Q}_i - \bar{Q}_j - \mathbb{E}\big[\bar{Q}_i - \bar{Q}_j \big] \geq \delta + \epsilon - \sum_{k \in Ch_i}\gamma^{H_k}\epsilon^U_k - \sum_{l \in Ch_j}\gamma^{H_l}\epsilon^U_l \big)~\label{error bound step} 
\end{align}
where the conditional dependence term is omitted after~\cref{proof start} for clarity.
The ~\cref{error step} follows from~\cref{lem: exp error} and the convergence condition.
The expected baseline estimator error $\epsilon^0$ is not generally known, though its magnitude can be limited by a term $\epsilon^U \geq |\epsilon^0|$.
Applying this inequality results in~\cref{error bound step}.
We use this term with~\cref{lem: cramer} to construct a Chernoff-Hoeffding bound on the sub-optimality as
\begin{align}
    P\big(Q_j - Q_i \geq \epsilon \mid \bar{Q}_i - \bar{Q}_j = \delta \big) 
    \leq & \exp{\Big(\frac{-2n_i n_j(|\delta + \epsilon - \zeta|^+)^2}{n_i\bar{\sigma}_i^2 + n_j\bar{\sigma}_j^2}\Big)}~\label{chernoff step}
\end{align}
where $\bar{\sigma}^2$ gives the variance of the value estimators.
Since the true variance is not generally known, we apply~\cref{lem: variance} with a union bound to arrive at
\begin{align}
    \leq & \exp{\Big(\frac{-2n_i n_j(|\delta + \epsilon - \zeta|^+)^2}{n_i\hat{\sigma}^{2, \alpha}_i + n_j\hat{\sigma}^{2, \alpha}_j}\Big)} + \alpha~\label{variance step 1}
\end{align}
which concludes the proof.
\end{proof}
\Cref{thm: gen ae bound} can be proved for action $a_i$ by applying the same procedure while omitting the terms related to $a_j$.

\subsection{Proof of~\Cref{thm: clt as bound}}
\begin{proof}
We begin this proof at~\cref{error bound step 1} of the general action error proof.
\begin{align}
    & P\big(\bar{Q}_i - \mathbb{E}\big[\bar{Q}_i \big] \geq \epsilon - \sum_{k \in Ch_i}\gamma^{H_k}\epsilon^U_k \big) \\
    \leq & 1 - \Phi(\epsilon - \zeta_i, 0, \frac{\bar{\sigma}^2}{n_i}) + \frac{C}{n_i}~\label{eq: CLT} \\
    \leq & 1 - \Phi(\epsilon - \zeta_i, 0, \frac{\hat{\sigma}^{2, \alpha}}{n_i}) + \alpha + \frac{C}{n_i}~\label{eq: CLT union}
\end{align}
The inequality of~\cref{eq: CLT} follows from the definition of the cumulative distribution function and application of~\cref{lem: berry-esseen}.
The inequality of~\cref{eq: CLT union} follows from application of the variance upper bound of~\cref{lem: variance} with the union bound, concluding the proof.
\end{proof}

\subsection{Proof of~\Cref{thm: clt ae bound}}
\begin{proof}
We begin this proof at~\cref{error bound step} of the general action error proof.
\begin{align}
    & P\big(\bar{Q}_i - \bar{Q}_j - \mathbb{E}\big[\bar{Q}_i - \bar{Q}_j \big] \geq \delta + \epsilon - \sum_{k \in Ch_i}\gamma^{H_k}\epsilon^U_k - \sum_{l \in Ch_j}\gamma^{H_l}\epsilon^U_l \big) \\
    \leq & 1 - \Phi(\delta + \epsilon - \zeta_i - \zeta_j, 0, \frac{\bar{\sigma}^2}{n_i + n_j}) + \frac{C}{n_i + n_j}~\label{eq: CLT 1} \\
    \leq & 1 - \Phi(\delta + \epsilon - \zeta_i - \zeta_j, 0, \frac{\hat{\sigma}^{2, \alpha}}{n_i + n_j}) + \alpha + \frac{C}{n_i + n_j}~\label{eq: CLT union 1}
\end{align}
The inequality of~\cref{eq: CLT 1} follows from the definition of the cumulative distribution function and application of~\cref{lem: berry-esseen}.
The inequality of~\cref{eq: CLT union 1} follows from application of the variance upper bound of~\cref{lem: variance} with the union bound, concluding the proof.
\end{proof}
\subsection{Proof of~\Cref{lem: exp error}}\label{sec: lem proof}
\begin{proof}
Define $\phi$ to be a policy that follows the Monte Carlo sampling strategy until a depth $h$ and then follows the optimal policy. 
Using this policy, we can define the expected error of an action value estimate to be
\begin{align}
    \mathbb{E}[\bar{Q}_i - Q_i] &= \mathbb{E}\Big[\frac{1}{n}\sum_{j \in Ch} \bar{v}_j - Q_i\Big] \\
    &= \mathbb{E}\Big[\frac{1}{n}\sum_{j \in Ch} \bar{v}_j - Q^\phi_i\Big] + Q^\phi_i - Q_i \\ 
    &= \frac{1}{n}\sum_{j \in Ch}\gamma^{H_j} \mathbb{E}\Big[(\hat{v}_j - v_j)\Big] + (Q^\phi_i - Q_i) \\
    &= \frac{1}{n}\sum_{j \in Ch} \gamma^{H_j} \epsilon^0_j + (Q^\phi_i - Q_i)\\
    &\leq \frac{1}{n}\sum_{j \in Ch} \gamma^{H_j} \epsilon^0_j~\label{err def}
\end{align}
where $v_j$ and $\hat{v}_j$ are the true and estimated values at the end of the sample trajectory, respectively. 
Intuitively, this is a decomposition of the expected error to the error of the baseline estimator and the expected error caused by the Monte Carlo sampling procedure.
The inequality in~\cref{err def} follows from the fact the the value the optimal policy value $Q_i \geq Q^\phi_i$ for any policy $\phi$.
Under the Monte Carlo convergence conditions, $\mathbb{E}[Q^\phi_i] = Q^\pi_i$.
\end{proof}

\subsection{Proof of~\Cref{lem: berry-esseen}}
\begin{proof}
We begin this proof with the statement of the Berry-Esseen theorem with known distributional parameters, with the third moment denoted $\rho = \mathbb{E}[|X|^3]$.
\begin{align}
    \sup_{x \in \mathbb{R}}|F_n(x) - \Phi(x)| & \leq \frac{C\rho}{\sqrt{n}\sigma^3} \\
    & = \Big( \frac{C}{\sqrt{n}} \Big) \Big( \frac{(1/n) \|X\|_3^3}{(1/n)^{3/2} \|X\|_2^3}\Big) \\
    & \leq \Big( \frac{C}{n} \Big)~\label{eq: norms}
\end{align}
The inequality in~\cref{eq: norms} follows from the fact that $\|X\|_p \geq \|X\|_q$ for any $p \leq q$, which results from H\"{o}lder's inequality~\citep{holder1889}.
\end{proof}

\end{document}